\documentclass[10pt,twocolumn,letterpaper]{article}

\usepackage[pagenumbers]{iccv} % To force page numbers, e.g. for an arXiv version

\usepackage{tikz}
\usetikzlibrary{positioning}
\usepackage{graphicx}
\usepackage{subcaption}
\usepackage{amsmath}
\usepackage{multirow}
\usepackage{adjustbox}
\usepackage{tabularx}

\usepackage{algorithm}
\usepackage{algpseudocode}
\usepackage{algorithmicx}
\usepackage{colortbl}
\usepackage{makecell}

\definecolor{iccvblue}{rgb}{0.21,0.49,0.74}
\usepackage[pagebackref,breaklinks,colorlinks,allcolors=iccvblue]{hyperref}

\title{SparseGS-W: Sparse-View 3D Gaussian Splatting in the Wild \\ with Generative Priors}

\author{
    Yiqing Li\textsuperscript{1}, 
    Xuan Wang\textsuperscript{2}, 
    Jiawei Wu\textsuperscript{1}, 
    Yikun Ma\textsuperscript{1}, 
    Zhi Jin\textsuperscript{1}\thanks{Corresponding author}
    \\
    \textsuperscript{1}Sun Yat-sen University \textsuperscript{2}Ant Research
}

\begin{document}
\maketitle
\begin{abstract}
Synthesizing novel views of large-scale scenes from unconstrained in-the-wild images is an important but challenging task in computer vision. Existing methods, which optimize per-image appearance and transient occlusion through implicit neural networks from dense training views (approximately 1000 images), struggle to perform effectively under sparse input conditions, resulting in noticeable artifacts. To this end, we propose SparseGS-W, a novel framework based on 3D Gaussian Splatting that enables the reconstruction of complex outdoor scenes and handles occlusions and appearance changes with as few as five training images. We leverage geometric priors and constrained diffusion priors to compensate for the lack of multi-view information from extremely sparse input. Specifically, we propose a plug-and-play Constrained Novel-View Enhancement module to iteratively improve the quality of rendered novel views during the Gaussian optimization process. Furthermore, we propose an Occlusion Handling module, which flexibly removes occlusions utilizing the inherent high-quality inpainting capability of constrained diffusion priors. Both modules are capable of extracting appearance features from any user-provided reference image, enabling flexible modeling of illumination-consistent scenes. Extensive experiments on the PhotoTourism and Tanks and Temples datasets demonstrate that SparseGS-W achieves state-of-the-art performance not only in full-reference metrics, but also in commonly used non-reference metrics such as FID, ClipIQA, and MUSIQ. 
\end{abstract}
\section{Introduction}
\label{sec:intro}
Novel view synthesis (NVS) from unconstrained photo collections is a long-standing challenge in the field of computer vision, with significant importance in areas such as virtual reality and augmented reality. Neural Radiance Field (NeRF)~\cite{mildenhall2021nerf} has made remarkable progress by significantly improving the quality and consistency of images rendered from arbitrary views. Recently, 3D Gaussian Splatting (3DGS)~\cite{kerbl20233DGS} has introduced an explicit point-based representation, achieving high-quality and real-time rendering. However, both methods are designed to model static scenes, lacking mechanisms for handling real-world datasets, such as large-scale Internet photo collections of tourist landmarks. These datasets usually consist of images captured at various times, from different viewpoints, under variable illumination, and often contain occlusions. 

To address the aforementioned problems, previous NeRF-based and 3DGS-based methods~\cite{martin2021nerf-W,chen2022haNerf,yang2023CRnerf,dahmani2024SWAG,kulhanek2024wildgaussians,xu2024wild-GS,wang2024WE-GS,zhang2024GS-W} incorporate neural networks to extract per-image appearance features and transient occlusions. However, above methods fundamentally rely on neural networks by leveraging the spatial consistency of static objects across densely captured views (approximately 1000 images), which makes it challenging to achieve high-quality rendering in extremely few-shot settings. As shown in~\cref{fig:teaser}(c), previous methods~\cite{zhang2024GS-W, kulhanek2024wildgaussians} produce noticeable elongated artifacts with only 5 input views.

The primary challenge lies in the coupling of multiple difficulties, including sparse inputs, transient occlusions, and dynamic appearance changes. In this work, we propose \textbf{SparseGS-W}, a novel framework to achieve high-quality and view-consistent scene reconstruction for sparse unconstrained image collections. 

Specifically, we first incorporate a pre-trained multi-view stereo model to obtain camera parameters and dense initial point cloud with rich geometric priors. To provide sufficient and reliable multi-view guidance for 3D Gaussians, we propose a plug-and-play \textit{Constrained Novel-View Enhancement} (CNVE) module, which utilizes constrained diffusion priors and attention injection strategy. The key insight of this module is to reformulate the problem of high-quality and view-consistent novel view synthesis as an iterative image enhancement task. Then, the enhanced views are iteratively exploited as the pseudo supervision for optimizing the 3D Gaussians. We propose an \textit{Occlusion Handling} (OH) module, which leverages the inherent high-quality inpainting capability of the constrained diffusion model to flexibly remove transient occlusions based on the user's preference. Both modules incorporate a simple yet effective AdaIn~\cite{huang2017adain} operation to flexibly extract appearance features from a given reference image, enabling efficient scene modeling without the need for an additional appearance extraction network. Additionally, we introduce a \textit{Progressive Sampling and Training Strategy} (PSTS) to improve the reconstruction of detail appearance in the scene.

Extensive validations on large-scale PhotoTourism datasets and Tanks and Temples datasets demonstrate that our method achieves state-of-the-art rendering quality and exhibits robust occlusion handling capabilities. Our contributions can be summarized as follows:
\begin{itemize}
    \item[$\bullet$] To the best of our knowledge, SparseGS-W is the first framework to handle few-shot novel view synthesis for unconstrained image collections.
    \item[$\bullet$] We propose a plug-and-play \textit{Constrained Novel-View Enhancement} module and an \textit{Occlusion Handling} module, both of which effectively leverage constrained diffusion priors for generating high-quality and view-consistent images free of transient occlusions.
    \item[$\bullet$] A \textit{Progressive Sampling and Training Strategy} to control the augmentation and sampling of novel views, as well as the training process of the Gaussian radiation field.
\end{itemize}
\section{Related Work}
\label{sec:Related Work}
%-------------------------------------------------------------------------
\subsection{Novel View Synthesis with Sparse Views}
NeRF~\cite{mildenhall2021nerf} and 3DGS~\cite{kerbl20233DGS}, as currently the most used novel view synthesis techniques, require a large number of training images to achieve good performance, which limits their practical applications. To address this, previous NeRFs~\cite{deng2022nerf_regulations1,kim2022nerf_regulations2,niemeyer2022nerf_regulations3,yang2023nerf_regulations4,yu2021nerf_regulations5,seo2023nerf_regulations6_flipnerf} introduce various regularizations to improve rendering quality with sparse input views. Other methods~\cite{deng2022nerf_regulations1,jain2021priors1,roessle2022priors2,wang2023priors3_sparsenerf,Wynn_2023_CVPR_priors4_diffusionnerf} attempt to introduce additional priors to extend NeRF, enabling similar performance.  With the development of 3DGS, some studies~\cite{li2024dngaussian,zhang2025cor-gs,paliwal2024coherentgs,zhu2025fsgs,liu20243dgs_enhancer} have explored using 3DGS as a replacement for NeRF to accomplish the few-shot novel view synthesis task. Notably, FSGS~\cite{zhu2025fsgs} introduces a Gaussian unpooling strategy and incorporates depth priors to guide the optimization of the Gaussian radiance field. CoherentGS~\cite{paliwal2024coherentgs} and CoR-GS~\cite{zhang2025cor-gs} apply different regularization techniques to constrain Gaussian points. DNGaussian~\cite{li2024dngaussian} further proposes depth regularization and normalization strategy based on depth priors to achieve more precise reconstruction results. However, these methods face challenges when applied to complex in-the-wild image collections, as they assume static scenes and do not consider transient occlusions and variable appearances.
%-------------------------------------------------------------------------
\subsection{Novel View Synthesis in the Wild}
To tackle in-the-wild scenes, NeRF-W~\cite{martin2021nerf-W} optimizes an appearance embedding for each image and trains a Multi-Layer Perceptron (MLP) to model a transient radiance field. Ha-NeRF~\cite{chen2022haNerf} and CR-NeRF~\cite{yang2023CRnerf} handle transient occlusions by optimizing an image-dependent 2D visibility map. They also introduce CNN-based appearance encoders to predict the global appearance of each image. More recently, GS-W~\cite{zhang2024GS-W} introduces intrinsic and dynamic appearance features to each Gaussian point to model various appearances. WildGaussians~\cite{kulhanek2024wildgaussians} extracts appearance embeddings from reference images with a learnable MLP and leverages DINO feature priors~\cite{oquab2024dinov2} to handle occlusions. Wild-GS~\cite{xu2024wild-GS} introduces an appearance decomposition strategy and an explicit modeling method to handle complex appearance variations. WE-GS~\cite{wang2024WE-GS} designs a lightweight spatial attention module to jointly predict transient masks and appearance embeddings. However, these methods utilize neural networks to model scenes based on the multi-view consistency of static objects, making it difficult to achieve good performance in the sparse-view setting.
% -----------------------------------------------------------------------
\subsection{Novel View Synthesis with Generative Priors}
Recently, leveraging generative priors for novel view synthesis has proven to be an effective approach. DreamFusion~\cite{pooledreamfusion} introduces Score Distillation Sampling (SDS) with a pre-trained diffusion model to guide 3D object generation. ProlificDreamer~\cite{wang2024prolificdreamer} presents variational score distillation, further improving the quality of 3D objects and effectively addressing issues in SDS, such as over-smoothing and over-saturation. Some methods~\cite{liu2023zero123,gao2024cat3d,qian2023magic123} achieve zero-shot NVS by incorporating 3D information into the diffusion prior. However, these methods require extensive training resources and struggle with real-world scenes. 3DGS-Enhancer~\cite{liu20243dgs_enhancer} fine-tunes a video diffusion model and designs a spatial-temporal decoder to achieve consistent NVS for unbounded outdoor scenes. However, it assumes scenes without occlusion or illumination variations and requires several days of training time, which limits its practical applicability to in-the-wild scenarios.
%-------------------------------------------------------------------------
\section{Preliminary}
\label{sec:preli}
%-------------------------------------------------------------------------
\subsection{3D Gaussian Splatting}
3D Gaussian Splatting~\cite{kerbl20233DGS} is a point-based method for explicitly representing 3D scenes. Each Gaussian has learnable attributes: center position \( \boldsymbol{\mu} \in \mathbb{R}^3 \), 3D covariance matrix \( \boldsymbol{\Sigma} \in \mathbb{R}^{3 \times 3} \), opacity \( o \in \mathbb{R} \) and color \( c \) represented by spherical harmonics. The influence of a 3D Gaussian on a point \( \boldsymbol{x} \) in the 3D world can be written as:
\begin{equation}
G(\boldsymbol{x}) = \frac{1}{(2\pi)^{3/2}|\boldsymbol{\Sigma}|^{1/2}} e^{-\frac{1}{2}(\boldsymbol{x}-\boldsymbol{\mu})^{T}\boldsymbol{\Sigma}^{-1}(\boldsymbol{x}-\boldsymbol{\mu})}.
\end{equation}
\indent To facilitate the optimization of the gradient of \(\boldsymbol{\Sigma}\) while preserving its physical properties, \(\boldsymbol{\Sigma}\) is decomposed into two learnable scaling matrix \( \boldsymbol{S} \) and rotation matrix \( \boldsymbol{R} \). For color rendering, 3D Gaussians are sorted by depth and projected into the 2D image plane, where the color of each pixel is rendered by a rasterizer through alpha-blending.
%-------------------------------------------------------------------------
\subsection{Diffusion Model}
Diffusion Models (DM)~\cite{ho2020ddpm,dhariwal2021diffusionbeatgan,rombach2022StableDiffusion} are generative models that progressively add random noise to transform a clean image \( \boldsymbol{x}_0 \sim p_{\text{data}}(\boldsymbol{x}) \) into Gaussian noise \( \boldsymbol{x}_T \sim \mathcal{N}(\mathbf{0}, \mathbf{I}) \) in the forward diffusion process:
\begin{equation}
    \boldsymbol{x}_t = \sqrt{\alpha_t} \cdot \boldsymbol{x}_0 + \sqrt{1 - \alpha_t} \cdot \boldsymbol{\epsilon}, \quad \text{where} \quad \boldsymbol{\epsilon} \sim \mathcal{N}(\mathbf{0}, \mathbf{I}).
\end{equation}

In the reverse diffusion process, a neural network \( \epsilon_\theta(\boldsymbol{x}_t, t) \) is trained to predict the added noise \( \boldsymbol{z}_{t-1} \) and progressively denoise \( \boldsymbol{x}_T \) until a clean image is obtained. This process can be formulated as:
\begin{equation}
    \boldsymbol{z}_{t-1} =  \epsilon_\theta(\boldsymbol{x}_t, t).
    \label{denoising}
\end{equation}

In this work, we leverage the prior knowledge of Stable Diffusion (SD)~\cite{rombach2022StableDiffusion}, which is built on a U-Net framework that includes residual blocks, cross-attention blocks, and self-attention blocks. In the self-attention block, the input intermediate features are query (\(\boldsymbol{Q}\)), key (\(\boldsymbol{K}\)), and value (\(\boldsymbol{V}\)). The output feature can be expressed as:
\begin{equation}
    \boldsymbol{F} = \text{softmax}\left(\frac{\boldsymbol{Q}\boldsymbol{K}^T}{\sqrt{d}}\right) \boldsymbol{V},
    \label{eq:selfattention}
\end{equation}
where \( d \) is the dimension of the embedding. Recent studies have shown that the attention mechanism controls spatial layout of the image~\cite{tumanyan2023plug_and_play, mou2024diffeditor}. We further use the query and key in the self-attention block to preserve the local structure of the image.
\begin{figure*}[ht!]
  \centering
    \includegraphics[width=1.0\linewidth]{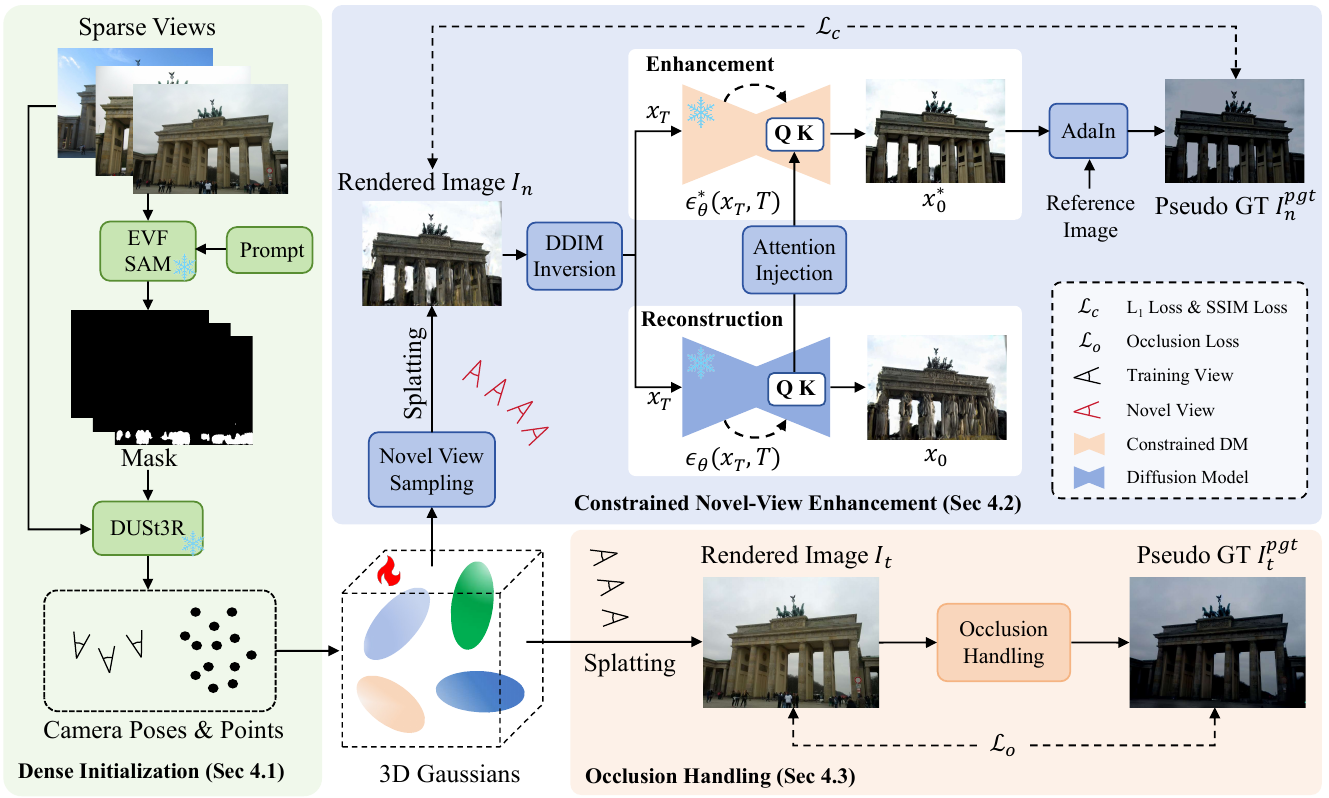}
    \vspace{-0.6cm}
   \caption{\textbf{An overviwe of the proposed SpraseGS-W framework.} Given unconstrained sparse images and user prompt, we perform dense initialization to obtain the initial point cloud, camera parameters, and occlusion masks. Then, we propose leveraging constrained diffusion priors to iteratively enhance the novel views rendered from the Gaussian radiance field and remove transient occluders.}
   \label{fig:pipeline}
   \vspace{-0.4cm}
\end{figure*}
% --------------------------------------------
\section{Method}
Given a sparse set of unconstrained images, our goal is to reconstruct the view-consistent, occlusion-free scene while allowing for appearance modification based on user's preference. In~\cref{subsec:4.1coarse}, we integrate a multi-view stereo model to generate dense 3D points and the corresponding camera poses. In~\cref{subsec:4.2enhancement}, we propose a plug-and-play \textit{Constrained Novel-View Enhancement} module to iteratively enhance the low-quality novel views. To handle transient objects, we propose an \textit{Occlusion Handling} module in~\cref{subsec:Occlusion Handling}. A \textit{Progressive Sampling and Training Strategy} is employed to progressively optimize 3D Gaussians in~\cref{subsec:4.4CL}. Finally, we describe the optimization process in~\cref{sec:opti}. An overview of our architecture is visualized as~\cref{fig:pipeline}. 

\subsection{Dense Initialization}
\label{subsec:4.1coarse}
We utilize DUSt3R~\cite{wang2024dust3r}, denoted as $\mathcal{G}_\theta$ to obtain dense 3D points with rich geometric priors and camera parameters. More specifically, given a sparse set of $N$ images $\mathcal{I} = \{\boldsymbol{I}_i^{gt}\}_{i=1}^N$, we can obtain points $\boldsymbol{P}$ and corresponding camera parameters $\mathcal{C} = \{\boldsymbol{C}_i\}_{i=1}^N$ as follows:
\begin{equation}
  \boldsymbol{P}, \mathcal{C} = \mathcal{G}_\theta(\mathcal{I}).
  \label{eq:dust3r}
\end{equation}
We observe that DUSt3R inherently assumes the absence of transient occlusions. This results in the generated points \( \boldsymbol{P} \) containing occlusion textures, which are incorrectly amplified by the diffusion model. To address this, we leverage a pre-trained segmentation model, EVF-SAM~\cite{zhang2024evf-sam}, to generate corresponding occlusion masks $\mathcal{M} = \{\boldsymbol{M}_i\}_{i=1}^N$ based on the text prompt by users. We then replace the colors of the occluded regions with Gaussian noise by: 
\begin{equation}
    \mathcal{I}_\mathcal{M} = \{ \boldsymbol{M}_i \odot \mathcal{N}(\boldsymbol{\mu}, \boldsymbol{\sigma}^2) + (\boldsymbol{1} - \boldsymbol{M}_i) \odot \boldsymbol{I}_i^{gt} \}_{i=1}^N,
\end{equation}
where \(\boldsymbol{1} \) is a same-shape matrix as \(\boldsymbol{M}_i \) but with all elements being 1, \(\mathcal{N}(\boldsymbol{\mu}, \boldsymbol{\sigma}^2)\) represents Gaussian noise with mean \(\boldsymbol{\mu}\) and variance \(\boldsymbol{\sigma}^2\) derived from non-masked regions and \(\odot\) denotes element-wise multiplication. We refine the~\cref{eq:dust3r} as follows:
\begin{equation}
  \boldsymbol{P}_\mathcal{M}, \mathcal{C} = \mathcal{G}_\theta(\mathcal{I}_\mathcal{M}).
\end{equation}

\subsection{Constrained Novel-View Enhancement}
\label{subsec:4.2enhancement}
We aim to enhance the low-quality images rendered from the Gaussian radiance field. To avoid content shift during enhancement, we use sparse training views as anchors to fine-tune the diffusion model, thus constraining the generative space to a clean subspace. To distinguish it from the original diffusion model \( \epsilon_\theta(\boldsymbol{x}_t, t) \), the constrained DM is denoted as \( \epsilon^*_\theta(\boldsymbol{x}_t, t) \).

For each rendered image \( \boldsymbol{I}_{n} \) from a sampled novel view, we apply DDIM Inversion~\cite{song2020ddiminversion} to convert it into a standard Gaussian distribution \( \boldsymbol{x}_T \). Subsequently, \( \boldsymbol{x}_T \) is input into two branches of reverse diffusion processes. The reconstruction process gradually denoises \( \boldsymbol{x}_T \) to reconstruct the original rendered image using \(\epsilon_\theta(\boldsymbol{x}_t, t)\), while the enhancement process focuses on generating the high-quality image \( \boldsymbol{x}_0^* \) using \(\epsilon^*_\theta(\boldsymbol{x}_t, t)\).

\begin{figure}[t]
    \includegraphics[width=1.0\linewidth]{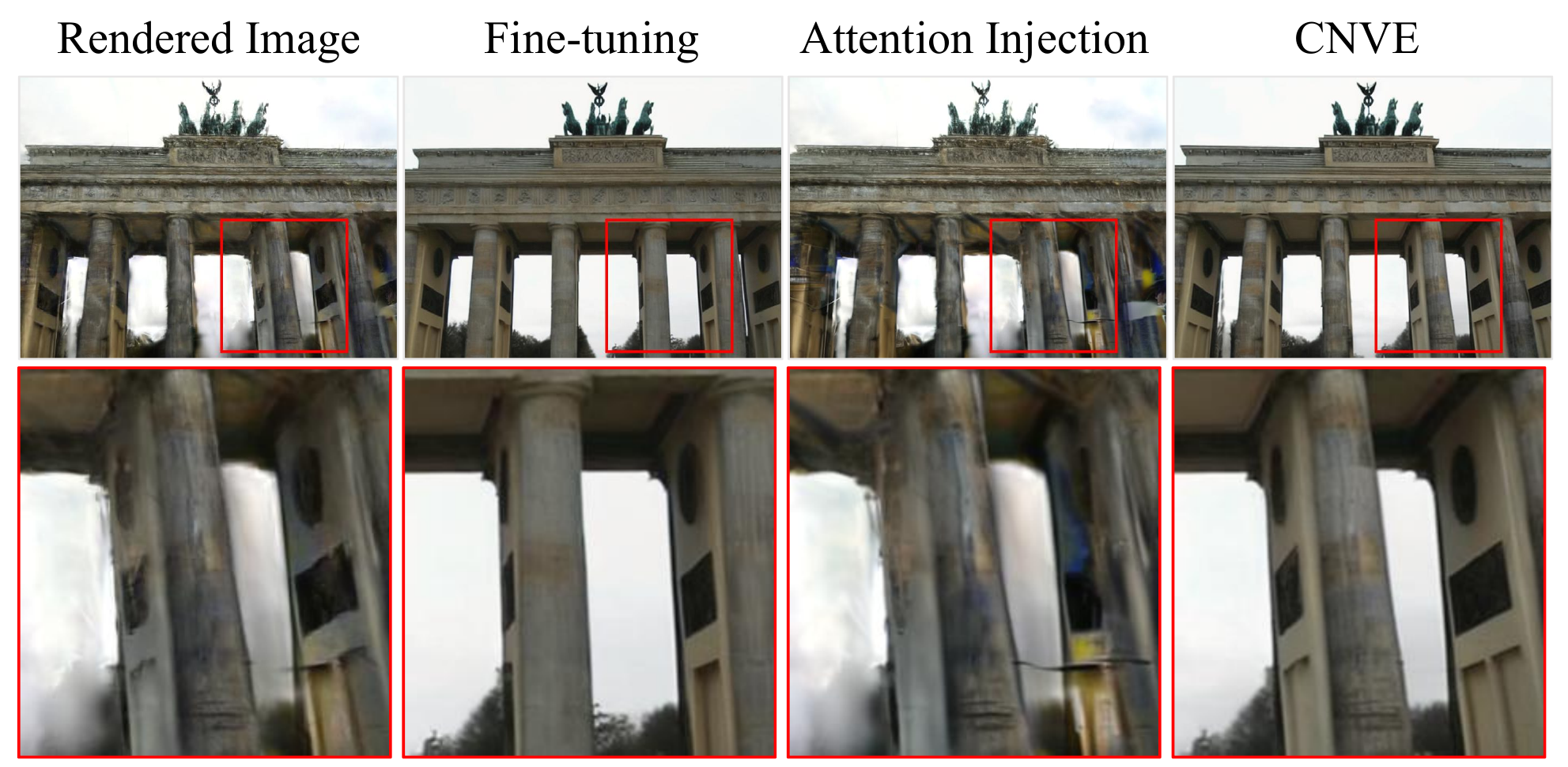}
    \vspace{-0.6cm}
   \caption{\textbf{Visualization results of CNVE module.} CNVE module can generate high-quality images by fine-tuning on the training views, but it struggles to preserve local image structure (the first pillar on the left in the zoom-in part). Injecting self-attention features helps maintain structure but cannot effectively remove artifacts. By combining these two strategies, CNVE achieves restoring high-fidelity, view-consistent novel views.}
   \label{fig:CPVE}
   \vspace{-0.3cm}
\end{figure}

As shown in~\cref{fig:CPVE}, we observe that although the constrained diffusion model can effectively restrict the generative space to produce high-quality novel views with the same content, it struggles to accurately preserve structure of the image, resulting in poor 3D consistency. To this end, we propose to inject structural features from the original rendered images into the enhancement process. Specifically, we use \( \boldsymbol{Q}_r \) and \( \boldsymbol{K}_r \) to denote the query and key from the self-attention block in the reconstruction process, while \( \boldsymbol{Q}_e \), \( \boldsymbol{K}_e \) and \( \boldsymbol{V}_e \) refer to the query, key and value from the enhancement process. The attention injection operation is performed by replacing \( \boldsymbol{Q}_e \) and \( \boldsymbol{K}_e \) with \( \boldsymbol{Q}_r \) and \( \boldsymbol{K}_r \) in~\cref{eq:selfattention}:  
\begin{equation}
    \boldsymbol{F}_e = \text{softmax}\left(\frac{\boldsymbol{Q}_r\boldsymbol{K}_r^T}{\sqrt{d}}\right) \boldsymbol{V}_e.
    \label{eq:attention1}
\end{equation}

\noindent Here, the output feature \( \boldsymbol{F}_e \) is utilized by the constrained \( \epsilon^*_\theta(\boldsymbol{x}_t, t) \) to predict noise during the enhancement process. After \( T \) steps of denoising, we obtain a high-quality image \( \boldsymbol{x}_0^* \) that preserves the content while eliminating artifacts. 

As it is challenging to model the appearance variation from sparse images, instead of training an extraction network, we utilize AdaIn~\cite{huang2017adain} to control the appearance of \( \boldsymbol{x}_0^* \) based on a user-provided reference image. The resulting image serves as the pseudo ground truth \( \boldsymbol{I}_{n}^{pgt} \) to supervise the original rendered image \( \boldsymbol{I}_n \), as defined in~\cref{eq:Lc}. Through this strategy, the appearance of the reference image is seamlessly integrated into the scene, achieving appearance consistency without additional training cost.
%-------------------------------------------------------------------------
\subsection{Occlusion Handling}
\label{subsec:Occlusion Handling}
% --------------------------------------------
\indent Prior works~\cite{chen2022haNerf,yang2023CRnerf,zhang2024GS-W,wang2024WE-GS,xu2024wild-GS} utilized neural networks to optimize image-dependent 2D visibility maps from dense training images. However, these methods face significant challenges when applied in a few-shot setting. Furthermore, their unsupervised manner lacks flexibility, making it difficult to remove specific occlusions based on user preference. To address these issues, we utilize a pre-trained segmentation network EVF-SAM~\cite{zhang2024evf-sam} to flexibly obtain occlusion masks \(\mathcal{M}\), as mentioned in~\cref{subsec:4.1coarse}. Subsequently, by treating the inpainting of occlusions as a masked image enhancement task, the proposed CNVE paradigm can naturally be applied to remove occlusions in the generation space. Specifically, given a rendered image \( \boldsymbol{I}_{t} \) from a training view and its ground truth \( \boldsymbol{I}_i^{gt} \), we use DDIM Inversion to obtain the corresponding latents \( \boldsymbol{x}_T^{\prime} \) and \( \boldsymbol{x}_T^{gt} \). The fused initial latent \( \boldsymbol{x}_T^* \) is computed via mask \( \boldsymbol{M}_i \):
\begin{equation}
    \boldsymbol{x}_T^* = \boldsymbol{M}_i \odot \boldsymbol{x}_T^{\prime} + (\boldsymbol{1} - \boldsymbol{M}_i) \odot \boldsymbol{x}_T^{gt}.
\end{equation}
\begin{figure*}[t]
    \centering
    \includegraphics[width=1.0\linewidth]{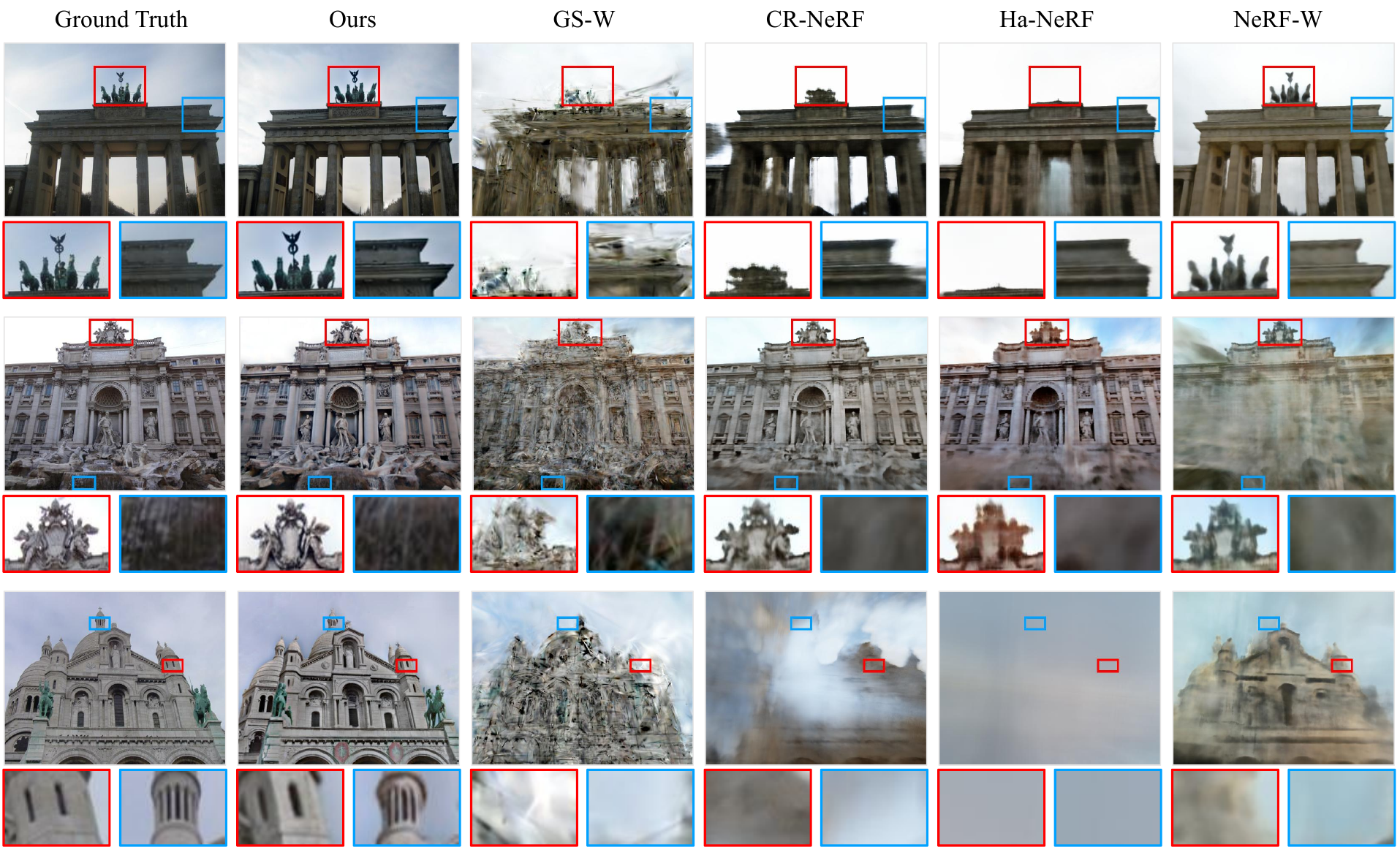}
    \vspace{-0.6cm}
    \caption{\textbf{Qualitative Comparison on PhotoTourism dataset.} Under the condition of sparse views, SparseGS-W is able to reconstruct more realistic and detailed scenes with less artifacts and blurring.}
    \label{fig:results}
    \vspace{-0.2cm}
\end{figure*}
\begin{table*}[h]
  \centering
    \begin{tabular}{>{\arraybackslash}p{32mm}>{\centering\arraybackslash}p{16mm}>{\centering\arraybackslash}p{14mm}>{\centering\arraybackslash}p{14mm}>{\centering\arraybackslash}p{12mm}c>{\centering\arraybackslash}p{12mm}>{\centering\arraybackslash}p{9mm}>{\centering\arraybackslash}p{9mm}}
    \toprule
    Method & PSNR↑ & SSIM↑ & LPIPS↓ & FID↓  & MUSIQ↑ & ClipIQA↑ & TT(m) & FPS \\
    \midrule
    NeRF-W~\cite{martin2021nerf-W} & 14.20  & 0.54  & 0.51  & 206  & 33.15  & 0.27  & 60 & \textless0.1 \\
    Ha-NeRF~\cite{chen2022haNerf} & 11.73  & 0.48  & \cellcolor{orange!20}0.38  & \cellcolor{orange!20}164  & 27.71  & 0.19  & 58 & \textless0.1 \\
    CR-NeRF~\cite{yang2023CRnerf} & \cellcolor{orange!20}15.08  & \cellcolor{red!20}0.59 & 0.47  & 202  & 36.72  & 0.23  & 39 & \textless0.1 \\
    3DGS~\cite{kerbl20233DGS}/3DGS\dag & 13.61/13.99 & 0.45/0.40 & 0.49/0.48 & 201/134  & 55.01/60.51  & 0.33/0.35 & 8/9 & 191  \\
    GS-W~\cite{zhang2024GS-W}/GS-W\dag  & 14.02/14.80  & 0.48/0.50  & 0.46/0.49  & 257/159  & 58.11/47.32  & 0.32/0.27  & 50/70 & 81  \\
    WildGS~\cite{kulhanek2024wildgaussians}/WildGS\dag & 11.78/14.73  & 0.39/0.41  & 0.51/0.46  & 269/119  & 47.55/\colorbox{orange!20}{60.70}  & 0.33/\colorbox{orange!20}{0.36}  & 420/603 & 117* \\
    \textbf{SparseGS-W\dag(Ours)}  & \cellcolor{red!20}19.01 & \cellcolor{orange!20}0.55  & \cellcolor{red!20}0.31 & \cellcolor{red!20}48 & \cellcolor{red!20}66.98 & \cellcolor{red!20}0.51 & 118 & 112  \\
    \bottomrule
    \end{tabular}%
    \vspace{-0.3cm}
      \caption{\textbf{Quantitative Comparison on PhotoTourism dataset for 5 input views.} We color each cell as \colorbox{red!20}{best} and \colorbox{orange!20}{second best}. \dag\; denotes using the same initial point cloud and camera poses from DUSt3R for fair comparisons, otherwise using sparse points and poses from COLMAP. WildGaussians is abbreviated as WildGS for brevity. * means number from the paper. TT denotes training time.}
  \label{tab:results}%
  \vspace{-0.4cm}
\end{table*}%
\indent Next, similar to the process outlined in~\cref{subsec:4.2enhancement}, we perform two branches of reverse diffusion processes for \( \boldsymbol{x}_T^* \). To distinguish symbols, we use a superscript to differentiate \(\boldsymbol{Q}_e^{\prime}\), \(\boldsymbol{K}_e^{\prime}\), \(\boldsymbol{V}_e^{\prime}\), \(\boldsymbol{Q}_r^{\prime}\), \(\boldsymbol{K}_r^{\prime}\) from \(\boldsymbol{Q}_e\), \(\boldsymbol{K}_e\), \(\boldsymbol{V}_e\), \(\boldsymbol{Q}_r\) and \(\boldsymbol{K}_r\). The attention injection process is formulated as follows:
\begin{equation}
\begin{aligned}
&\boldsymbol{Q}_{OH}  = \boldsymbol{M}_i \odot \boldsymbol{Q}_e^{\prime} + (\boldsymbol{1} - \boldsymbol{M}_i) \odot \boldsymbol{Q}_r^{\prime}, \\
&\boldsymbol{K}_{OH}  = \boldsymbol{M}_i \odot \boldsymbol{K}_e^{\prime} + (\boldsymbol{1} - \boldsymbol{M}_i) \odot \boldsymbol{K}_r^{\prime}, \\ 
&\boldsymbol{F}_e^{\prime} = \text{softmax}\left(\frac{\boldsymbol{Q}_{OH}(\boldsymbol{K}_{OH})^T}{\sqrt{d}}\right) \boldsymbol{V}_e^{\prime}.
\end{aligned}
\end{equation}
\indent After \(T\) steps of denoising followed by AdaIn, we obtain a high-fidelity and occullsion-free image that serves as pseudo ground truth \( \boldsymbol{I}_t^{pgt} \). The weighted fusion and injection operations help minimize inherent information loss during the diffusion process, ensuring the constrained SD focuses on the masked regions. 
%-------------------------------------------------------------------------
\subsection{Progressive Sampling and Training Strategy}
\label{subsec:4.4CL}
\noindent\textbf{Novel View Sampling.}\quad To prevent overfitting to sparse training views, we employ novel view augmentation:
\begin{equation}
\begin{aligned}
&\mathcal{C}^{\prime} = \bigcup_{1 \leq i < j \leq N} \text{SLERP}(\boldsymbol{C}_i, \boldsymbol{C}_j, \alpha), \quad \alpha \sim \mathcal{U}(0,1), \\
&\mathcal{P}^{\prime} = \bigcup_{\boldsymbol{C}_k \in \mathcal{C}} (\boldsymbol{t}_k + \boldsymbol{\epsilon}, \boldsymbol{q}), \quad \boldsymbol{\epsilon} \sim \mathcal{N}(\boldsymbol{0}, \boldsymbol{\delta}), \\
&\mathcal{C}^* = \mathcal{C}^{\prime} \cup \mathcal{P}^{\prime}.
\end{aligned}
\label{eq:sample}
\end{equation}
\noindent Here, SLERP means spherical linear interpolation. The $\mathcal{P}^{\prime}$ is inspired by previous work~\cite{zhu2025fsgs}. $\boldsymbol{t}_k \in \boldsymbol{C}_k$ denotes camera locations, $\boldsymbol{q}$ is a quaternion representing the rotation averaged from the two closest training views.

\noindent\textbf{Progressive Sampling and Training.}\quad We adapt a progressive training strategy that transitions from simple to complex views. The process begins by randomly selecting training views to optimize the Gaussians for $\tau_c$ iterations. Subsequently, we gradually integrate sampled novel views $\mathcal{C}^*$ into training process in three stages based on their complexity. These novel views are categorized into three levels: simple, medium, and difficult based on their Euclidean distance from the training views. To balance the trade-off between errors and useful priors introduced by the diffusion model, we sample $\mathcal{C}^*$ for training with a probability \(\beta\). The PSTS ultimately results in a multi-view consistent Gaussian radiance field. For further details, please refer to the supplementary material.
%-------------------------------------------------------------------------
\subsection{Optimization}
\label{sec:opti}
Following 3DGS~\cite{kerbl20233DGS}, we use the L1 loss \( \mathcal{L}_1 \) and SSIM loss~\cite{wang2004ssim} \( \mathcal{L}_{SSIM} \) to measure the difference between the rendered image and the ground truth (GT) image \( \boldsymbol{I}_i^{gt} \). For the rendered training view \( \boldsymbol{I}_t\), when the iteration is less than \(\tau_o\), we use the corresponding mask \(\boldsymbol{M}_i\) to mask out transient occlusions. Once the iteration exceeds \(\tau_o\), we employ its pseudo GT \( \boldsymbol{I}_t^{pgt}\) to supervise the entire image. Let \(\overline{\boldsymbol{M}_i}=\boldsymbol{1}-\boldsymbol{M}_i \). The loss function can be written as~\cref{eq:Lo}. For the rendered novel view \( \boldsymbol{I}_n\), we use its pseudo GT \( \boldsymbol{I}_n^{pgt}\) as the supervision mentioned in~\cref{eq:Lc}. The total loss function is formulated as in~\cref{eq:total_L}, where \(\lambda_1\), \(\lambda_{2}\) and \(\lambda_3\) are 0.8, 0.2, 1.0, respectively. 
\begin{equation}
\mathcal{L}_o = 
\begin{cases}
\mathcal{L}_c(\overline{\boldsymbol{M}_i} \odot \boldsymbol{I}_t, \overline{\boldsymbol{M}_i} \odot \boldsymbol{I}_i^{gt}), & \text{if iteration} < \tau_o; \\
\mathcal{L}_c(\boldsymbol{I}_t, \boldsymbol{I}_t^{pgt}), & \text{if iteration} \ge \tau_o.
\end{cases}
\label{eq:Lo}
\end{equation}
% --------------------------------
\begin{equation}
\mathcal{L}_c(\boldsymbol{I}_n, \boldsymbol{I}_n^{pgt}) = \lambda_1 \mathcal{L}_1(\boldsymbol{I}_n, \boldsymbol{I}_n^{pgt}) + \lambda_{2} \mathcal{L}_{SSIM}(\boldsymbol{I}_n, \boldsymbol{I}_n^{pgt}).
\label{eq:Lc}
\end{equation}
% --------------------------------
\begin{equation}
    \mathcal{L} = \mathcal{L}_o + \lambda_3 \mathcal{L}_c.
    \label{eq:total_L}
\end{equation}
\vspace{-0.6cm}
\begin{figure*}[htbp!]
    \centering
    \includegraphics[width=1.0\linewidth]{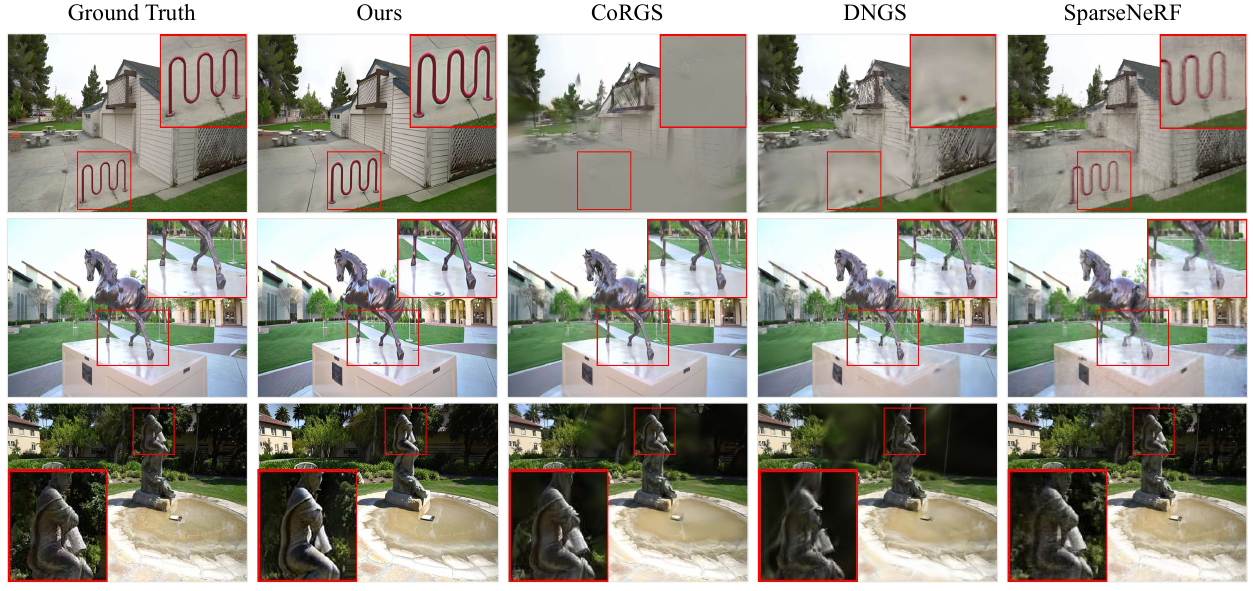}
    \vspace{-0.6cm}
    \caption{\textbf{Qualitative Comparison on Tanks and Temples dataset.} Our method outperforms other baselines in the task of few-shot NVS.}
    \label{fig:results_tanks}
    \vspace{-0.2cm}
\end{figure*}
% ---------------------------------------------------------------------
\begin{table*}[htbp]
  \centering
    \begin{tabular}{>{\arraybackslash}p{30mm}>{\centering\arraybackslash}p{16mm}>{\centering\arraybackslash}p{14mm}>{\centering\arraybackslash}p{14mm}>{\centering\arraybackslash}p{12mm}c>{\centering\arraybackslash}p{12mm}>{\centering\arraybackslash}p{9mm}>{\centering\arraybackslash}p{9mm}}
    \toprule
    Method & PSNR↑ & SSIM↑ & LPIPS↓ & FID↓  & MUSIQ↑ & ClipIQA↑ & TT(m) & FPS \\
    \midrule
    SparseNeRF~\cite{wang2023priors3_sparsenerf} & 19.13 & 0.55 & 0.44 & 191 & 44.66 & 0.25 & 73 & \textless0.1 \\
    3DGS~\cite{kerbl20233DGS}/3DGS\dag & 16.01/16.99 & 0.43/0.55 & 0.42/0.36 & 182/200 & 64.03/63.59 & 0.38/0.41 & 5/7 & 317/291 \\
    FSGS~\cite{zhu2025fsgs}/FSGS\dag & 17.17/18.11 & 0.53/0.52 & 0.42/0.34 & 205/126 & 59.24/62.51 & 0.28/0.33 & 20/21 & 297/170 \\
    DNGS~\cite{li2024DNGS}/DNGS\dag & 18.32/19.30 & 0.56/0.56 & 0.43/0.32 & 216/105 & 57.46/66.13 & 0.30/\colorbox{orange!20}{0.46} & 7/8 & 286/79 \\
    CoRGS~\cite{zhang2025cor-gs}/CoRGS\dag & 19.03/\colorbox{orange!20}{19.92} & 0.59/\colorbox{orange!20}{0.63} & 0.37/\colorbox{orange!20}{0.27} & 269/\colorbox{orange!20}{72} & 65.05/\colorbox{orange!20}{68.11} & 0.29/0.41 & 7/14 & 238/149 \\
    \textbf{Sparse-GS\dag(Ours)} & \cellcolor{red!20}21.58 & \cellcolor{red!20}0.68 & \cellcolor{red!20}0.22 & \cellcolor{red!20}55 & \cellcolor{red!20}69.80 & \cellcolor{red!20}0.50 & 28 & 102 \\
    \bottomrule
    \end{tabular}%
    \vspace{-0.3cm}
    \caption{\textbf{Quantitative Comparison on Tanks and Temples dataset for 3 input views.} We color each cell as \colorbox{red!20}{best} and \colorbox{orange!20}{second best}. \dag\; denotes using the same initial point cloud and camera poses from DUSt3R for fair comparisons, otherwise using sparse points and poses from COLMAP. TT represents the training time, measured in minutes.}
    \label{tab:tanks}
    \vspace{-0.3cm}
\end{table*}%
% ----------------------------------------------------------------
\section{Experiments}
\subsection{Experimental Setup}
\noindent\textbf{Dataset.}\quad Since the few-shot novel view synthesis on large-scale \textbf{PhotoTourism datasets}~\cite{jin2021phototourism} is a novel and more challenging task, we do not utilize the exact same test set as previous methods based on dense training views. For each scene, we select 10 images from various viewpoints as the test set and select 5 images as the training set for all methods. Notably, we not only follow previous works~\cite{yang2023CRnerf,chen2022haNerf,zhang2024GS-W} by using the same three scenes: Brandenburg Gate, Sacre Coeur, and Trevi Fountain, but also include additional scenes: Palace of Westminster and Pantheon Exterior. To further validate the robustness of our method, we compare it with several sparse-view methods on the \textbf{Tanks and Temples dataset}. For each scene, we sample 3 views from every 8-frame clip as the training set and test on the remaining 1/8 of the images at a resolution of 688 \(\times\) 512.

\noindent\textbf{Metrics and baselines.}\quad We report quantitative results of the full-reference metrics PSNR, SSIM~\cite{wang2004ssim}, and LPIPS~\cite{zhang2018lpips}. We observe that the results of SSIM do not align with the visual quality of the image, a phenomenon also mentioned in the field of image restoration~\cite{blau2018IR1,jinjin2020IR2,gu2022IR3,yu2024ScaleUp_restoration}. Therefore, to provide a more comprehensive evaluation, we also report the non-reference metrics FID~\cite{heusel2017fid}, ClipIQA~\cite{wang2023clipiqa}, and MUSIQ~\cite{ke2021musiqa} for all methods. We evaluate our method against the SOTA open-source in-the-wild methods NeRF-W~\cite{martin2021nerf-W}, Ha-NeRF~\cite{chen2022haNerf}, CR-NeRF~\cite{yang2023CRnerf}, 3DGS~\cite{kerbl20233DGS}, GS-W~\cite{zhang2024GS-W}, and WildGaussians~\cite{kulhanek2024wildgaussians}. Additionally, we include comparisons with several SOTA few-shot methods SparseNeRF~\cite{wang2023priors3_sparsenerf}, FSGS~\cite{zhu2025fsgs}, DNGS~\cite{li2024dngaussian}, and CoRGS~\cite{zhang2025cor-gs}.

\noindent\textbf{Implementation Details.}\quad We implement our method using PyTorch~\cite{paszke2019pytorch}. To constrain the generative space, we fine-tune Stable Diffusion V1.5 using DreamBooth~\cite{ruiz2023dreambooth} for 400 iterations with a batch size of 1 and a learning rate of \(2\times10^{-6}\), which takes less than 2 minutes. For PhotoTourism datasets, we use the \textbf{prompt ``tourist''} to generate masks $\mathcal{M}$ predicted by EVF-SAM~\cite{zhang2024evf-sam} and train our model on a single NVIDIA RTX 4090 GPU for a total of 7,500 iterations. Specifically, to ensure stability, we first train on the training views for $\tau_c = \text{5,500}$ iteration, followed by 2,000 iterations applying the CNVE module with \(\beta\) set to 0.3. The OH module is applied at the \(\tau_o=\text{6,500}\) iteration. For Tanks and Temples dataset, we train the model with 1,000 iterations and the CNVE module is applied after 500 iterations. The OH module is not utilized in this case, as there are no occlusions present in this dataset. We follow other hyperparameter settings as 3DGS~\cite{kerbl20233DGS}, except without using adaptive density control. We set the DDIM sampling and inversion steps \( T \) to 50.
\begin{figure*}[t!]
    \centering
    \includegraphics[width=1.0\linewidth]{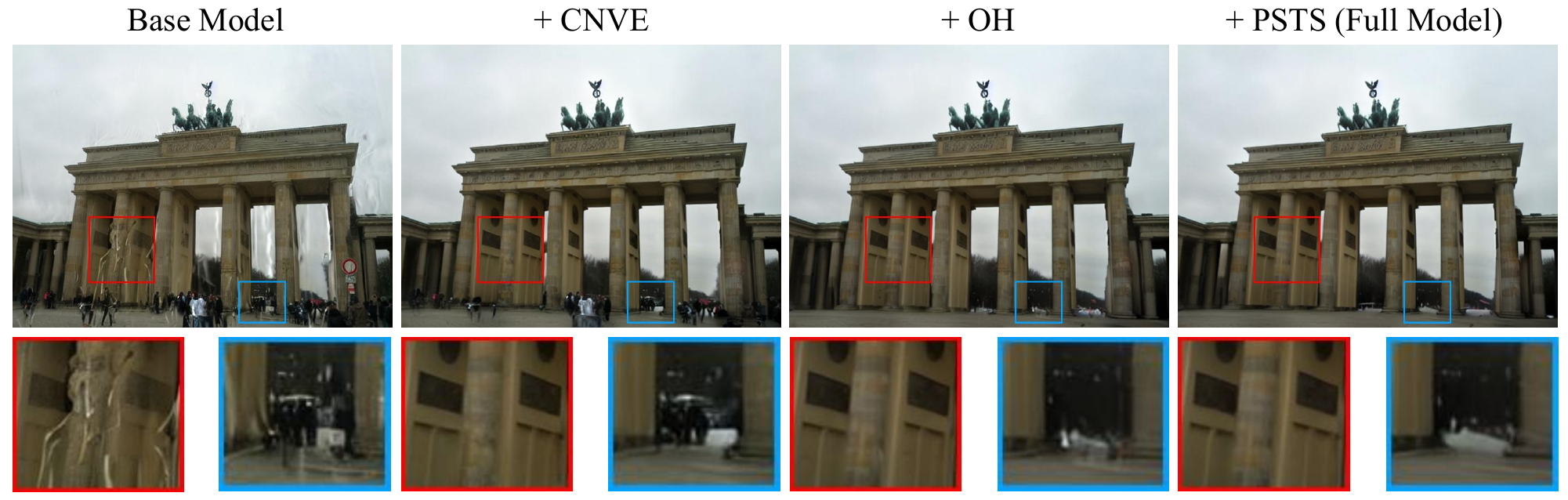}
    \vspace{-0.6cm}
    \caption{\textbf{Qualitative results of ablation studies.} With the proposed CNVE, OH modules and the PSTS, SparseGS-W effectively eliminates transient occlusions and removes numerous artifacts, significantly enhancing the results in the few-shot scenarios.}
    \label{fig:ablation}
    \vspace{-0.4cm}
\end{figure*}
\subsection{Comparison}
\noindent\textbf{PhotoTourism.}\quad Quantitative results and visualizations are shown in~\cref{tab:results} and~\cref{fig:results}. Our method demonstrates significant performance improvements over state-of-the-art methods. Compared to the representative NeRF-based method CR-NeRF~\cite{yang2023CRnerf}, our method achieves notable enhancements, with PSNR, LPIPS, and FID improved by $+25.7\%$, $+34.0\%$ and $+76.2\%$, respectively. Compared to the representative 3DGS-based methods GS-W~\cite{zhang2024GS-W} with the same initial points and camera poses, our method improves the PSNR, LPIPS and FID by $+28.1\%$, $+36.7\%$ and $+69.8\%$, respectively. In particular, comparisons with GS-W and WildGaussians under the same point cloud and camera parameters demonstrate the effectiveness of our framework. Our method provides additional multi-view supervision, which guides the optimization of Gaussians and represents illumination changes.

\noindent\textbf{Tanks and Temples.}\quad To evaluate the robustness of our framework, we conduct experiments on the Tanks and Temples dataset, which does not contain dynamic appearance changes and transient occlusions. Quantitative results and visualizations are presented in~\cref{tab:tanks} and~\cref{fig:results_tanks}. Our method not only achieves the highest scores across all full-reference and non-reference metrics but also demonstrates superior visual quality. These results show that the framework we designed for sparse large-scale Internet photo collections has robustness to expand to more general scenarios.
%-------------------------------------------------------------------------
\begin{table}[t!]
  \centering
  \setlength{\tabcolsep}{1.3mm}{
  \resizebox{\columnwidth}{!}{
    \begin{tabular}{ccccccc}
    \toprule
    \multicolumn{1}{c}{} & PSNR↑ & SSIM↑ & LPIPS↓ & FID↓  & MUSIQ↑ & ClipIQA↑ \\
    \midrule
    Base Model & 15.38  & 0.48  & 0.43  & 116  & 59.99  & 0.31  \\
     + CNVE & 18.42  & 0.52  & 0.36 & 110  & 62.63  & 0.46  \\
     {\makecell{+ OH}} & 18.70 & 0.54 & 0.32 & \cellcolor{red!20}48 & 66.84 & 0.47  \\
     {\makecell{+ PSTS (Full)}} & \cellcolor{red!20}19.01  & \cellcolor{red!20}0.55  & \cellcolor{red!20}0.31  & \cellcolor{red!20}48 & \cellcolor{red!20}66.98 & \cellcolor{red!20}0.51  \\
    \bottomrule
    \end{tabular}%
    }
    % \vspace{-0.2cm}
    \caption{\textbf{Ablation study on PhotoTourism dataset.}}
  \label{tab:ablation}  }
  \vspace{-0.4cm}
\end{table}%
%--------------------------------------------------------------------
\subsection{Ablation studies}
In~\cref{tab:ablation} and~\cref{fig:ablation}, we ablate our method on PhotoTourism 5-view setting. Base Model extends 3DGS by integrating DUSt3R. From the zoomed-in details in the red box, we observe that the \textit{Constrained Novel-View Enhancement} (CNVE) module significantly reduces artifacts, improves rendering quality, and reshapes scene geometry. After adding the \textit{Occlusion Handling} (OH) module, pedestrians are effectively removed from the scene. The zoomed-in details in the blue box demonstrate the effectiveness of the \textit{Progressive Sampling and Training Strategy} (PSTS) in improving high-frequency details within the scene.
% ---------------------------------------------------------------
\subsection{Plug-and-play CNVE experiment}
\begin{figure}[t!]
    \centering
    \includegraphics[width=1.0\linewidth]{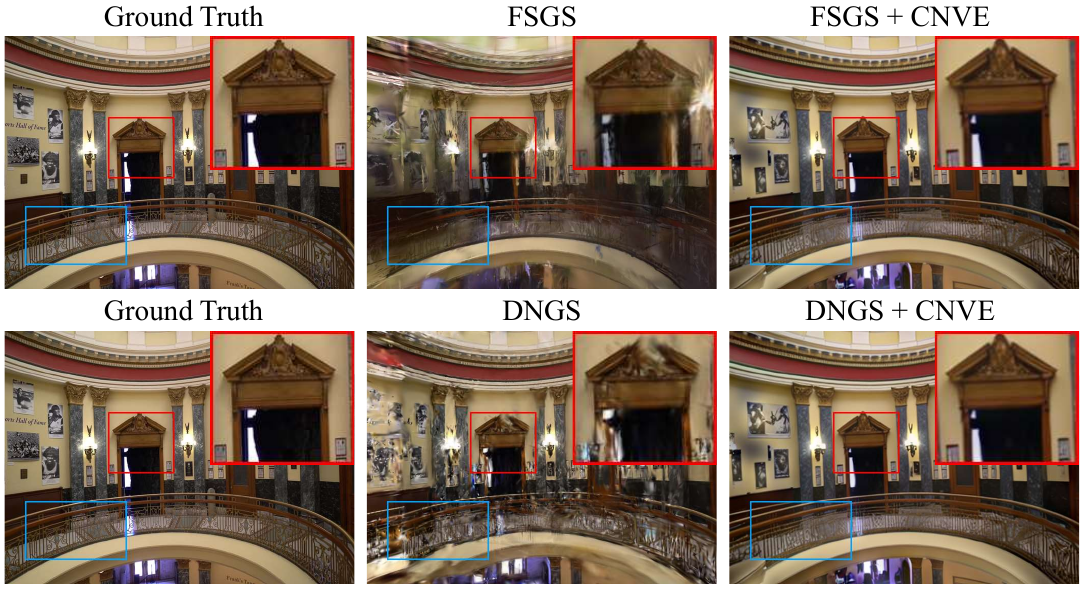}
    \vspace{-0.6cm}
    \caption{\textbf{Visualization Results of plug-and-play CNVE.}}
    \label{fig:CNVE_FSGS_DNGS}
    \vspace{-0.2cm}
\end{figure}

We validate our plug-and-play CNVE module on Tanks and Temples under the 3-view setting. Visualization and quantitative results are reported in~\cref{fig:CNVE_FSGS_DNGS} and~\cref{tab:CNVE}. The results show that the CNVE module effectively improves the rendering quality and recovers complete geometric structures with only a modest increase in training time.
\begin{table}[t!]
  \centering
  \setlength{\tabcolsep}{1.3mm}{
  \resizebox{\columnwidth}{!}{
    \begin{tabular}{lccccccc}
    \toprule
    Method & PSNR↑ & SSIM↑ & LPIPS↓ & FID↓  & MUSIQ↑ & ClipIQA↑ & TT(m) \\
    \midrule
    FSGS~\cite{zhu2025fsgs} & 17.17  & 0.53  & 0.42  & 205  & 59.24  & 0.28 & 20  \\
     FSGS+CNVE & \cellcolor{red!20}19.28  & \cellcolor{red!20}0.57  & \cellcolor{red!20}0.34 & \cellcolor{red!20}114 & \cellcolor{red!20}68.01 & \cellcolor{red!20}0.51 & 60  \\
     \midrule
     DNGS~\cite{li2024DNGS} & 18.32  & \cellcolor{red!20}0.56  & 0.43  & 216  & 57.46  & 0.30 & 7 \\
     DNGS+CNVE & \cellcolor{red!20}19.01  & 0.54  & \cellcolor{red!20}0.40  & \cellcolor{red!20}132  & \cellcolor{red!20}60.73  & \cellcolor{red!20}0.39 & 39  \\
    \bottomrule
    \end{tabular}%
    }
    \vspace{-0.2cm}
    \caption{\textbf{Quantitative results of plug-and-play CNVE.}}
  \label{tab:CNVE}  }
  \vspace{-0.6cm}
\end{table}%
\section{Conclusion}
In this paper, we propose SparseGS-W, a novel framework for reconstructing unconstrained in-the-wild scenes with as few as five training images by leveraging constrained diffusion priors. We believe that our method is a step forward towards real-world applications. However, estimating the camera poses in the sparse-view setting is a challenging problem. Our method struggles with poor initial camera poses. We leave it for future work.
{
    \small
    \bibliographystyle{ieeenat_fullname}
    \bibliography{main}
}

\end{document}